\documentclass[acmsmall]{acmart}
\AtBeginDocument{%
  }

\usepackage{tabularx}
\usepackage{multirow}
\usepackage[table]{xcolor}
\usepackage{algorithm}
\usepackage{algorithmic} 
\usepackage{amsmath}   

\usepackage{amssymb}     
\acmJournal{JACM}
\acmVolume{37}
\acmNumber{4}
\acmArticle{111}
\acmMonth{8}

\begin{document}

%%
%% The "title" command has an optional parameter,
%% allowing the author to define a "short title" to be used in page headers.
\title{Dual-branch Prompting for Multimodal Machine Translation}

%%
%% The "author" command and its associated commands are used to define
%% the authors and their affiliations.
%% Of note is the shared affiliation of the first two authors, and the
%% "authornote" and "authornotemark" commands
%% used to denote shared contribution to the research.
\author{Jie Wang}
\email{jackwang@swjtu.edu.cn}
\affiliation{%
  \institution{School of Computing and Artificial Intelligence, Southwest Jiaotong University}
  \city{Chengdu}
  \country{China}
}
\orcid{0000-0001-8690-8164}

\author{Zhendong Yang}
\email{Menta180016@163.com}
\affiliation{%
  \institution{School of Computer and Software Engineering, Xihua University}
  \city{Chengdu}
  \state{610039}
  \country{China}
}

\author{Liansong Zong}
\email{lszong@my.swjtu.edu.cn}
\affiliation{%
  \institution{School of Computer and Software Engineering, Xihua University}
  \city{Chengdu}
  \country{China}
}
\affiliation{%
  \institution{School of Computing and Artificial Intelligence, Southwest Jiaotong University}
  \city{Chengdu}
  \country{China}
}

\author{Xiaobo Zhang}
\affiliation{%
  \institution{School of Computing and Artificial Intelligence, Southwest Jiaotong University}
  \city{Chengdu}
  \country{China}
}

\author{Dexian Wang}
\affiliation{%
  \institution{School of Intelligent Medicine, Chengdu University of Traditional Chinese Medicine}
  \city{Chengdu}
  \country{China}
}
\email{wangdexian@cdutcm.edu.cn}

\author{Ji Zhang}
\authornote{Ji Zhang is the corresponding author.}
\affiliation{%
  \institution{School of Computing and Artificial Intelligence, Southwest Jiaotong University}
  \city{Chengdu}
  \country{China}
}
\email{jizhang@swjtu.edu.cn}

\renewcommand{\shortauthors}{Wang et al.}

%%
%% The abstract is a short summary of the work to be presented in the
%% article.
\begin{abstract}
Multimodal Machine Translation (MMT) typically enhances text-only translation by incorporating aligned visual features. Despite the remarkable progress, state-of-the-art MMT approaches often rely on paired image-text inputs at inference and are sensitive to irrelevant visual noise, which limits their robustness and practical applicability. To address these issues, we propose D$^2$P-MMT, a diffusion-based dual-branch prompting framework for robust vision-guided translation. Specifically, D$^2$P-MMT requires only the source text and a reconstructed image generated by a pre-trained diffusion model, which naturally filters out distracting visual details while preserving semantic cues. During training, the model jointly learns from both authentic and reconstructed images using a dual-branch prompting strategy, encouraging rich cross-modal interactions. To bridge the modality gap and mitigate training-inference discrepancies, we introduce a distributional alignment loss that enforces consistency between the output distributions of the two branches. Extensive experiments on the Multi30K dataset demonstrate that D$^2$P-MMT achieves superior translation performance compared to existing state-of-the-art approaches. Our code is publicly available at \url{https://github.com/MentaY/DDP}.
\end{abstract}

%%
%% The code below is generated by the tool at http://dl.acm.org/ccs.cfm.
%% Please copy and paste the code instead of the example below.
%%


%%
%% Keywords. The author(s) should pick words that accurately describe
%% the work being presented. Separate the keywords with commas.

% \received{20 February 2007}
% \received[revised]{12 March 2009}
% \received[accepted]{5 June 2009}

%%
%% This command processes the author and affiliation and title
%% information and builds the first part of the formatted document.
\maketitle
\section{Introduction}
Over the years, Neural Machine Translation (NMT) has evolved from statistical methods to sophisticated deep neural networks based on the Transformer architecture \cite{vaswani2017attention}, establishing itself as the standard paradigm in the field. While recent NMT models have achieved impressive benchmarks, they predominantly rely on textual data, often lacking the rich contextual cues inherent in real-world environments \cite{gao2024multimodal}. Consequently, similar to other multimodal tasks~\cite{wang2025diagllm, wang2024cross, zhangji2023channel, zhangji2025closer}, there has been a growing interest in Multimodal Machine Translation (MMT), a setting that enhances text-only systems by incorporating shared visual perceptions of objects and scenes ~\cite{li2022vision}. This integration aims to provide visual grounding to the translation task, thereby alleviating semantic ambiguity where text-only NMT systems often struggle. Therefore, MMT research holds great significance for developing more robust, accurate, and contextually-aware translation systems that better reflect real-world communication \cite{fei2023scene}.

For the incorporation of visual signals, previous works have predominantly employed specifically engineered encoder-decoder architectures designed to jointly process linguistic and visual inputs~\cite{lin2020dynamic}, or focused on extracting specific object-level embeddings~\cite{zhao2021word}. However, these methods typically necessitate paired images during inference, which limits their practical applicability in scenarios where visual input is unavailable. Towards this end, recent research has attempted to eliminate this dependency by either learning hallucination networks to produce pseudo-visual features \cite{li2022valhalla,fei2023scene,long2021generative}, utilizing generative models to synthesize images from text~\cite{fang2022neural}, or adopting retrieval modules to fetch relevant images from auxiliary datasets \cite{zhang2020neural, feng2023mkvse}.

\begin{figure}[htbp]
	\centering %
	\includegraphics[width=1\columnwidth]{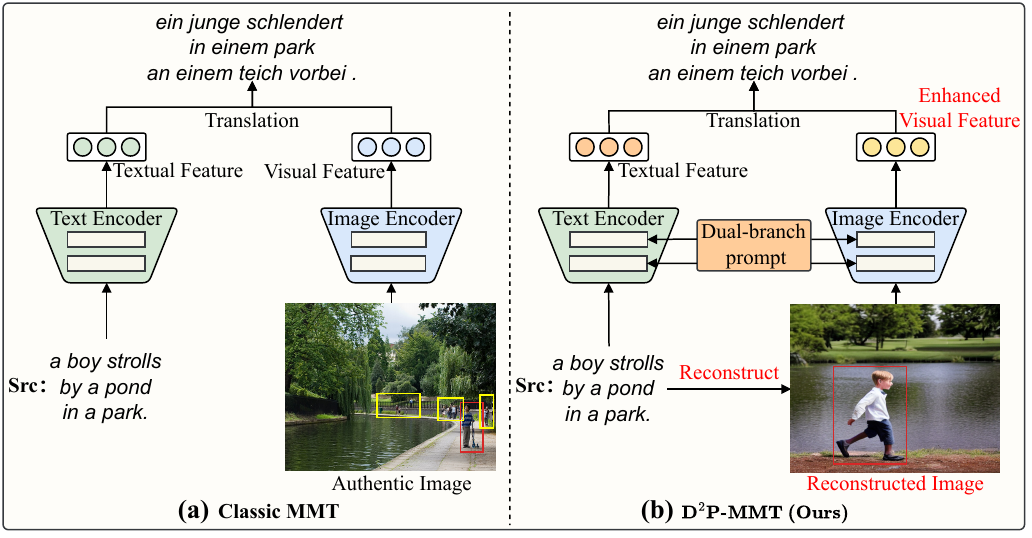}
	\caption{Illustrations of classic MMT model and our proposed D$^2$P-MMT framework. In the authentic image, the red bounding box highlights the main content of the sentence, while the yellow bounding box indicates redundant information. In D$^2$P-MMT, irrelevant visual information is filtered out by reconstructing the image.}
	\label{Figure 1}
\end{figure}

Despite these advancements, existing approaches face a challenge in balancing robustness with effective visual utilization. A major limitation lies in the distributional gap between authentic visual data used during training and the synthetic representations employed at inference. Furthermore, even when authentic images are available, they frequently contain substantial background information that is extraneous to the core semantic content. As illustrated in Fig~\ref{Figure 1} (a), the authentic image paired with the source sentence often includes distracting details (highlighted in yellow) that diverge from the text (highlighted in red). Current MMT models tend to be sensitive to such irrelevant visual noise, which hampers their ability to disentangle informative cues from distracting content \cite{zhang2025reliable}. In contrast, Fig~\ref{Figure 1} (b) conceptually illustrates the objective of our work: refining visual representations to filter out noise while preserving semantic consistency.

In this work, we aim to overcome these limitations and enhance the robustness of vision-guided translation by proposing \textbf{D$^{2}$P-MMT}, a diffusion-based dual-branch prompting framework. Instead of relying on raw visual inputs that may contain noise, we employ a pre-trained Stable Diffusion model to generate textually grounded reconstructed images. This process naturally filters out distracting visual details while preserving essential semantic cues. Specifically, we design a dual-branch prompting strategy that allows the model to jointly learn from both authentic and reconstructed images during training. In the visual branch, we introduce staged prompt modules to guide the learning of textual prompts within the visual context. To bridge the modality gap and mitigate the discrepancy between training and inference, we incorporate a Kullback-Leibler (KL) divergence loss that enforces consistency between the output distributions of the two branches. This tactfully encourages the model to learn robust representations that generalize across both visual modalities, allowing us to discard authentic images at test time.

The main contributions of our work can be summarized as follows:
\begin{itemize}
    \item We propose D$^{2}$P-MMT, a novel dual-branch prompting framework that leverages diffusion-based reconstructed images to enhance multimodal machine translation while eliminating the need for authentic images at inference time.
    \item We design a dual-branch prompting strategy that enables joint learning from both authentic and reconstructed visual inputs. We build a cross-branch coupling function to explicitly bridge the visual and textual modalities, facilitating robust joint training and improving generalization capability.
    \item We conduct extensive experiments on the Multi30K benchmark, demonstrating that D$^{2}$P-MMT achieves significant improvements over strong baselines.
\end{itemize}

\section{Related work}
\subsection{Machine Translation}

Machine Translation (MT) has been examined through various lenses \cite{bhadwal2020machine, rajan2009rule, singh2020corpus, rahman2018corpus, laskar2019neural, xu2020dual}, witnessing a paradigm shift from early Rule-Based Machine Translation (RBMT) and Statistical Machine Translation (SMT) to the current state-of-the-art Neural Machine Translation (NMT). As discussed in earlier literature, RBMT approaches typically relied on explicit linguistic rules. Bhadwal et al. \cite{bhadwal2020machine} and Rajan et al. \cite{rajan2009rule} utilized bilingual dictionaries alongside associated grammatical, semantic, and morphological rules to transform source language structures into the target. In contrast, SMT methods shifted focus to data-driven approaches. Singh et al. \cite{singh2020corpus} and Rahman et al. \cite{rahman2018corpus} employed corpus-based techniques, such as n-gram models and phrase extraction from bilingual corpora, to identify optimal translation matches.

Finally, NMT revolutionized the field by employing end-to-end deep learning models to capture complex linguistic relationships. With respect to architectural evolution, Saini et al. \cite{laskar2019neural} utilized a bidirectional encoder-decoder structure based on LSTMs to connect hidden layers from both directions, significantly improving translation accuracy. Further building upon this, recent works have explored data augmentation strategies, such as generating pseudo-sentence pairs from monolingual corpora \cite{xu2020dual}, to advance text-based translation. However, despite these advancements, the performance of pure NMT systems remains limited in contexts where semantics are inherently ambiguous or sparse.

\subsection{Multimodal Machine Translation}

Multimodal Machine Translation (MMT) has emerged as a prominent area of research, aiming to enhance neural machine translation by incorporating auxiliary visual information, such as images or videos. The core premise is that visual context can augment linguistic representations, thereby improving the model's ability to disambiguate complex scenes \cite{li2022vision}.

Historically, MMT research has witnessed a paradigm shift, evolving from early RNN-based encoder-decoder architectures to the now-dominant Transformer-based models leveraging sophisticated attention mechanisms \cite{yao2020multimodal}. To further generate pseudo-representations of visual features, Fei et al.~\cite{fei2023scene} utilized graph convolutional networks to encode visual scene graphs.
With the advent of attention mechanisms, the focus shifted towards more explicit alignment and fusion strategies. Nishihara et al. \cite{nishihara2020supervised} proposed a supervised visual attention mechanism to explicitly align textual tokens with corresponding image regions. Inspired by these advancements, researchers began utilizing attention mechanisms to fuse and align visual object embeddings \cite{li2023multimodality,helcl2018cuni}. Specifically, Calixto et al.~\cite{calixto2017doubly} introduced a doubly-attentive module to independently attend to source words and visual features.

Subsequent approaches have sought to refine these cross-modal interactions through more complex structures. Wang and Xiong \cite{wang2021efficient} introduced strategies to mask irrelevant objects in the visual modality. Additionally, Yin et al. \cite{yin2020novel} adopted a unified multimodal graph to capture semantic interactions, while Lin et al. \cite{lin2020dynamic} proposed a dynamic context-guided module using capsule networks to address insufficient fusion guidance.
Recent advances have facilitated the development of frameworks such as denoising-driven visual representation learning \cite{wu2024pixel, song2025pushing, song2022improving} and prior/prompt-guided learning \cite{zhang2021deep, wang2025visual}, which perform multimodal modeling with a robustness-oriented and prompt-oriented paradigm. Specifically, Song et al. \cite{song2025uni} unified visual modeling to optimize the adaptability and semantic alignment efficiency of visual information, thereby enhancing the robustness of cross-modal modeling. Furthermore, Lu et al. \cite{lu2025visual}, Liu et al. \cite{liu2025condition}, and Wen et al. \cite{wen2025all} proposed adaptive self-prompting strategies for effective visual restoration and prediction tasks. 

However, these methods primarily target unimodal vision restoration or recognition tasks. In contrast, our approach adapts this robust prompting paradigm to the multimodal domain. Specifically, we focus on enhancing robustness in multimodal machine translation by utilizing reconstructed text-aligned visual representations and introducing a unique dual-branch prompting framework, which facilitates deep cross-modal interaction to filter out irrelevant visual noise.

\section{Method}
% This section commences by outlining the problem formulation (Section~\ref{function_1}). Subsequently, we introduce the key steps of our proposed D$^2$P-MMT: Image Feature Reconstruction (Section~\ref{function_2}), Dual-Branch Prompting Learning(Section~\ref{function_3}), and Consistency Training (Section~\ref{function_4}). Finally, the language translation process is detailed in Section~\ref{function_5}. D$^2$P-MMT’s overall structure is shown in Fig.~\ref{Figure 2}.

\subsection{Preliminaries}
\label{function_1}
\noindent
\textbf{Multimodal Machine Translation.} 
The MMT task is formally defined as utilizing auxiliary visual signals (e.g., images or videos) to enhance textual translation. MMT systems typically operate within an encoder-decoder framework. Given a source sentence $x=(x_{1}, \ldots,x_{N})$ of length $N$ and a target sentence $y = (y_{1}, \ldots, y_{M})$ of length $M$, the model estimates the conditional probability of the target sequence. A Transformer-based architecture comprises an encoder $f_{T}^{Enc}$ and a decoder $f_{T}^{Dec}$. Specifically, an image encoder $f_{V}^{Enc}$ maps the image $v$ to a visual latent representation $z$. This representation is integrated with the word embeddings of $x$ and fed into the decoder $f_{T}^{Dec}$, serving as visual conditioning to guide the generation process. The conditional probability is formulated as:
\begin{equation}
\begin{aligned}
p(y|x,z;\mathrm{f_{T}}) &= \prod_{i=1}^{n} p(y_{i}|y_{<i},x,z) \\
&= \prod_{i=1}^{n} \mathbf{f_{T}^{Dec}}(y_{i}\mid y_{<i},\mathbf{f_{T}^{Enc}}(x;\theta_{e});\theta_{d},z),
\end{aligned}
\end{equation}
where $z = f_{V}^{Enc}(v)$, and $\mathrm{f_{T}} = (f_{T}^{Enc}, f_{T}^{Dec})$. Parameters $\theta_{e}$ and $\theta_{d}$ correspond to the encoder and decoder, respectively. The decoder $f_{T}^{Dec}$ predicts the probability of each token $y_i$ using cascaded attention mechanisms, attending to both the encoder output $f_{T}^{Enc}(x)$ and the preceding target tokens $y_{<i}$. The model is optimized by minimizing the cross-entropy loss on a triplet dataset $(x, v, y)$:
\begin{equation}
\ell_T(\mathbf{f_T};z)=\mathbb{E}_{(x,z,y)}[-\log p(y\mid x,z;\mathbf{f_T})].
\end{equation}

\noindent
\textbf{Multimodal Transformer.} 
We adopt the Multimodal Transformer \cite{yao2020multimodal} as our base architecture. It extends the standard Transformer \cite{vaswani2017attention} by replacing the single-modal self-attention layer in the encoder with a cross-modal self-attention layer, enabling the simultaneous processing of text and visual inputs. Let the source language sequence be $H^{x} = \{h_{1}^{x}, \ldots, h_{N}^{x}\}$ and the image representation be $H^{i} = \{h_{1}^{i}, \ldots, h_{K}^{i}\}$, where $N$ is the source sentence length and $K$ is the number of image features. In the multimodal self-attention layer, text and visual representations are concatenated to form joint query vectors:
\begin{equation}\mathbf{H_{xv}}=\begin{bmatrix}\mathbf{H}^x;\mathbf{H}^iW^i\end{bmatrix},\end{equation}
where $W^i$ is a learnable weight matrix. The text representations $H^x$ serve as key and value vectors. The output of the multimodal self-attention layer is calculated as:
\begin{equation}\mathbf{A}_i=\sum_{j=1}^N\tilde{\alpha}_{ij} \cdot \left(W_V \cdot h_j^x\right),\end{equation}
where $\tilde{\alpha}_{ij}$ are the self-attention scores computed by the softmax function:
\begin{equation}\tilde{\alpha}_{ij}=\text{softmax}\left(\frac{(W_Q \cdot \tilde{h}_i)(W_K \cdot \tilde{h}_j^x)^\mathrm{T}}{\sqrt{d_k}}\right), \end{equation}
where $W_{Q}$, $W_{K}$, $W_{V}$ are learnable weights, and $d_{k}$ is the dimension of the key vectors. The final output of the Multimodal Transformer Encoder is subsequently fed into $f_{T}^{Dec}$ to generate the target translation.

\noindent
\textbf{Latent Diffusion Model.} Latent Diffusion Model (LDM) \cite{rombach2022high} is a text-to-image generation model based on the diffusion process, which models data distribution in a low-dimensional latent space to generate high-quality images. Given the high accuracy requirements for machine translation in MMT tasks, we adopted the Stable Diffusion model, which is based on the latent diffusion model. The model is composed of a VAE, a U-Net, and a CLIP text encoder. 

The training of Stable Diffusion involves a forward diffusion process and a reverse denoising process. In the forward phase, the VAE encoder projects the input image from pixel space into a low-dimensional latent space to obtain a latent representation, to which Gaussian noise is progressively added. In the reverse phase, a U-Net with shared parameters is iteratively applied to remove noise. Crucially, the U-Net is conditioned on the text description via the CLIP text encoder, integrating semantic information through cross-attention. Finally, the VAE decoder reconstructs the image from the denoised latent representation. We utilize a pre-trained Stable Diffusion model to generate reconstructed images of the source sentences.

\subsection{Approach Overview}

\begin{figure}[htbp]
\centering
\includegraphics[width=1\textwidth]{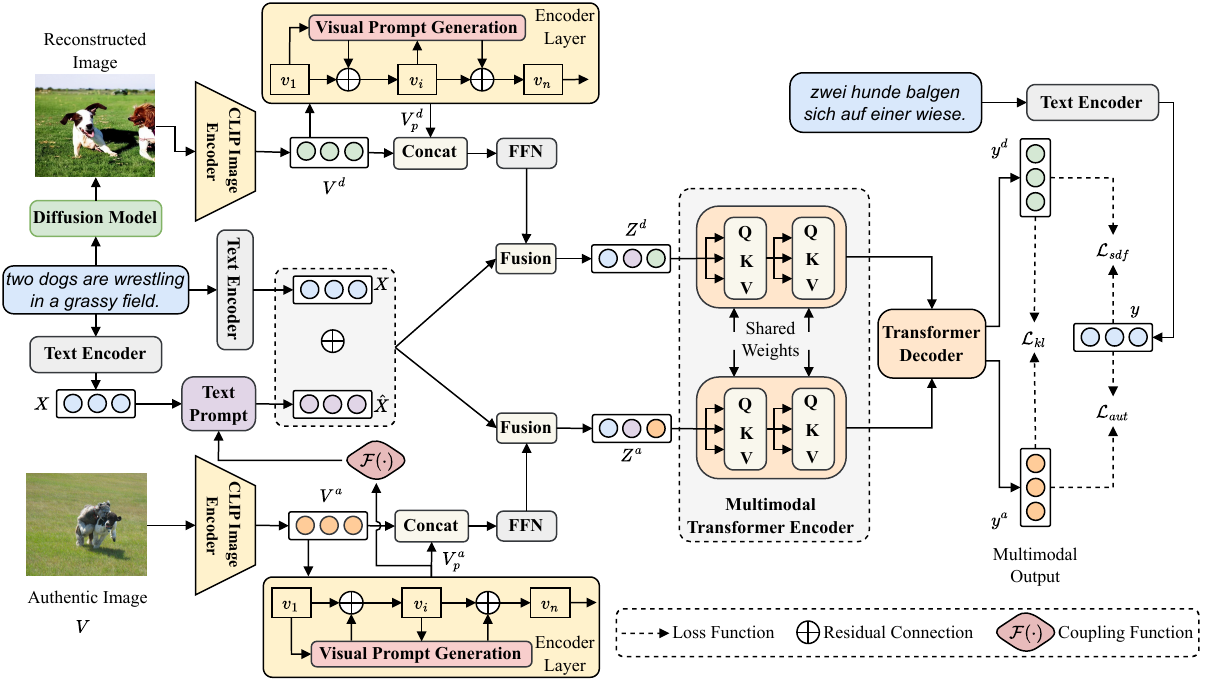} 
\caption{The overall framework of the proposed D$^{2}$P-MMT model. It consists of four stages: \textbf{image feature reconstruction}, \textbf{visual prompt generation}, \textbf{dual-branch prompting}, and \textbf{language translation}. Images are reconstructed using pretrained diffusion models, and text prompts are adjusted based on visual prompts via a coupling function ${\mathcal{F}(\cdot)}$ to facilitate cross-modal interaction. The final translation output is derived from two input streams: the reconstructed fused representation $Z^{d}$ and the authentic fused representation $Z^{a}$.
}
\label{Figure 2}
\end{figure}

The overall framework of the proposed D$^{2}$P-MMT is illustrated in Fig.~\ref{Figure 2}. The core objective of our approach is to effectively mitigate the interference of visual noise by incorporating semantically reconstructed images. To this end, we propose a dual-branch prompting strategy designed to facilitate multi-level knowledge extraction from both textual and visual modalities, thereby fostering robust cross-modal interaction.

Specifically, the proposed pipeline comprises four integral stages: image feature reconstruction, dual-branch prompt learning, consistency training, and the final translation process. Initially, given a source sentence $x$ and its corresponding authentic image $v$, we utilize a pre-trained Stable Diffusion model to generate a reconstructed image for each source sentence, serving to augment the input representation. The MMT model is subsequently trained jointly using both authentic and reconstructed visual inputs. 

During the training phase, the model operates via two parallel information streams: one processing the reconstructed image (upper branch in Fig.~\ref{Figure 2}) and the other processing the authentic image (lower branch). In the visual branch, prompt blocks are introduced to refine visual representations. Simultaneously, in the textual branch, visual prompts condition the text prompts via a cross-branch coupling function, establishing a synergistic interaction mechanism between the two modalities. These dual representations are fed into a Multimodal Transformer, yielding two distinct output distributions, $y^{d}$ and $y^{a}$. Both branches are optimized via negative log-likelihood loss against the ground truth target $y$. Furthermore, to bridge the gap between the two visual modalities, we introduce a consistency loss that enforces alignment between the predictions of the reconstructed and authentic branches. This constraint is crucial for ensuring the reliability of the reconstructed module during inference. Consequently, during the inference stage, the model generates translations conditioned solely on the text input $x$ and the generated reconstructed image, effectively eliminating the dependency on authentic visual data.

\subsection{Image Feature Reconstruction}
\label{function_2}
We aim to leverage the generative capabilities of the Stable Diffusion model to reconstruct visually grounded images from textual descriptions. This Stable Diffusion model operates on the latent space of a VAE, employing a forward diffusion process to inject noise and a reverse process to reconstruct data samples. The objective function of the latent diffusion model is defined as:
\begin{equation}
\mathcal{L}_{\mathrm{LDM}}=\mathbb{E}_{\varepsilon(v),y,\epsilon\sim\mathcal{N}(0,1),t}\bigl[\parallel\epsilon-\epsilon_\theta\bigl(z_t,t,\tau_\theta(y)\bigr)\parallel_2^2\bigr],
\end{equation}
where $\epsilon$ represents Gaussian noise, $\varepsilon$ denotes the VAE encoder, $\epsilon_{\theta}$ and $\tau_{\theta}$ represent the U-Net and text encoder respectively, and $\theta$ denotes the model parameters. $y$ and $v$ correspond to the input text and image, while $z_{t}$ represents the latent representation at time step $t$.

In our framework, the SD model guides the image reconstruction process conditioned on source sentence descriptions. Text embeddings $c$ are extracted using the CLIP ViT-L/14 text encoder based on the input text $x$. The denoising process is then conditioned on these embeddings to generate latent representations:
\begin{equation}
\begin{aligned}
v_{t-1}&=\frac{1}{\sqrt{\alpha_t}} \cdot \Big(v_t-\sqrt{1-\alpha_t} \cdot \epsilon_\lambda(v_t,t,c)\Big)+\sqrt{1-\alpha_{t-1}} \cdot \epsilon_\lambda(v_t,t,c),
\end{aligned}
\end{equation}
where $\epsilon_{\lambda}(v_{t},t,c)$ is the noise predicted by the model, $\alpha_t$ represents the noise schedule parameter, and $c$ represents the text description embedding vector. Finally, the VAE decoder projects the denoised representations from the latent space to the pixel space to yield the reconstructed image, which serves as a text-aligned visual input for downstream multimodal translation.

\subsection{Dual-Branch Prompt Learning}
\label{function_3}

The core objective of our framework is to enable the model to capture deeper semantic nuances from the source language by leveraging visual signals, particularly for ambiguous or underspecified expressions. In contrast to mainstream unimodal enhancement approaches, our dual-branch prompting strategy facilitates a dynamic adaptation of both textual and visual representation spaces. This mechanism allows the model to extract not only rich linguistic semantics but also highly relevant visual context. Specifically, we integrate multi-stage prompt blocks into the hierarchical layers of the visual branch. These enhanced visual features serve a dual purpose: they are fused with subsequent feature maps for visual refinement and, crucially, are utilized to guide the generation of textual prompts, thereby fostering a deep, synergistic interaction between the two modalities.

\subsubsection{Visual Prompt Generation} 
As demonstrated in prior work \cite{islam2021broad}, hierarchical visual feature representations significantly contribute to network generalization. High-level features encode global scene semantics, while lower-level features preserve fine-grained spatial details. To fully exploit semantic information across multiple scales and enrich the representation space, we embed prompt blocks into multiple layers of the CLIP visual encoder. 

\begin{figure}[htbp]
	\centering 
	\includegraphics[width=0.7\columnwidth]{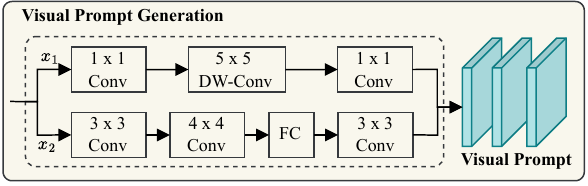}
	\caption{Implementation of visual multi-level prompt enhancement module.}
	\label{Figure 3}
\end{figure}

As illustrated in Fig.~\ref{Figure 3}, the input visual features are split along the channel dimension and processed by two parallel branches. The global branch aims to capture coarse-grained semantic structures and long-range dependencies, whereas the local branch focuses on modeling detailed spatial patterns and fine-level visual information. The outputs of both branches are then fused to form a unified visual prompt.

Given the input image features $x\in \mathbb{R}^{B \times L \times D}$ (where $B$ is the batch size, and $L$ and $D$ represent the length and dimension, respectively), we perform dimensional expansion and channel grouping. The input is split into $x_{1}\in \mathbb{R}^{B \times \frac{C}{2} \times L \times D}$ and $x_{2}\in \mathbb{R}^{B \times \frac{C}{2} \times L \times D}$. 

The global branch compresses channel information and applies depth-wise convolution to enhance spatial modeling capability, enabling the extraction of high-level semantic context:
\begin{equation}
\begin{aligned}
\hat{x_{1}}&=P_\text{global}\\
&=P_{\text{Conv2d}_{1 \times 1}}(P_{\text{Conv2d}_{5 \times 5}}(P_{\text{Conv2d}_{1 \times 1}}(x_{1}))) 
\end{aligned}
\end{equation}  
where $\hat{x_{1}}\in \mathbb{R}^{B \times C \times L \times D}$, $P_{Conv2d}$ represents a convolutional layer, and the subscripts denote the kernel size.

In contrast, the local branch focuses on capturing fine-grained visual details through stacked convolutions and a fully connected layer:
\begin{equation}
\begin{aligned}
\hat{x}_2&=P_\text{local}\\
&=P_{\text{Conv2d}_{3\times3}}\left(\text{Linear}\left(P_{\text{Conv2d}_{4\times4}}\left(P_{\text{Conv2d}_{3\times3}}(x_2)\right)\right)\right)
\end{aligned}
\end{equation}  

Additionally, a ReLU activation function is applied after each convolutional layer to introduce non-linearity, facilitating the learning of complex features while ensuring non-negativity.

Finally, $x_{1}$ and $\hat{x_{1}}$ are connected via residual connections, and a linear layer maps them to the same space as $\hat{x_{2}}$. The results are concatenated along the channel dimension to generate the complete visual prompt $V_{p}$:
\begin{equation}
\begin{aligned}
V_{p}&=VPG(x)\\
&=\text{Concat}\left(\text{Linear}\big(P_\text{global}(x_{1})+x_{1}\big),P_\text{local}(x_{2})\right)\end{aligned}
\end{equation} 
The VPG module effectively leverages both holistic semantics and fine-grained details through these parallel branches to enhance the expressiveness of visual representations.

\begin{algorithm}[h]
\caption{Visual-Guided Language Prompt Generation}
\label{alg1}
\begin{algorithmic}[1]
\renewcommand{\algorithmicrequire}{\textbf{Input:}}
\renewcommand{\algorithmicensure}{\textbf{Output:}}
\REQUIRE Source sentence embedding $X \in \mathbb{R}^{L \times d_{w}}$, Visual prompt $V_{p} \in \mathbb{R}^{L \times d_{v}}$, Projection function $\mathcal{F}(\cdot)$, Learnable weight matrix $W$.
\ENSURE The visual-guided language prompt $X_{p}$.

\STATE \textit{Step 1: Dimensional Alignment}
\STATE Initialize mapping of visual features to language space:
\STATE $\tilde{V}_{p} \leftarrow \mathcal{F}(V_{p})$ \quad \textit{// Bridge gap from $d_v$ to $d_l$}

\STATE \textit{Step 2: Cross-Modal Attention Setup}
\STATE Set Query, Key, and Value matrices:
\STATE $Q_{x} \leftarrow X\cdot W_{Q}$
\STATE $K_{v} \leftarrow \tilde{V}_{p}\cdot W_{K}$
\STATE $V_{v} \leftarrow \tilde{V}_{p}\cdot W_{V}$

\STATE \textit{Step 3: Compute Attention Weights}
\STATE Calculate similarity scores between textual query and visual key:
\STATE $Scores \leftarrow \left(\frac{X \cdot W_{Q} \cdot \left(\tilde{V}_{p} \cdot W_{K}\right)^{\top}}{\sqrt{d}}\right)$
\STATE Apply Softmax normalization to obtain attention weights $\alpha$:
\STATE $\alpha \leftarrow \text{Softmax}(Scores)$

\STATE \textit{Step 4: Visual-Language Fusion}
\STATE Generate final prompt with learnable transformation:
\STATE $X_{p} \leftarrow \alpha \cdot V_{v}$

\RETURN $X_{p}$
\end{algorithmic}
\end{algorithm}

\subsubsection{Visual Language Prompt Integration}
\label{function_vl}
Effective multimodal learning requires simultaneous adjustment of visual and language branches to achieve contextual optimization. A naive approach, termed `independent prompting', adjusts the visual prompt $V_{p}$ and language prompt $X_{p}$ independently. Although this satisfies prompt completeness, it lacks explicit collaboration between modalities. To address this limitation, we have designed a Dual-Branch Prompting method to supersede independent strategies by enabling visual features to explicitly guide the learning of language context. This mechanism first bridges the dimensional discrepancy between the visual and linguistic latent spaces via a coupling function, ensuring synergy between visual and text prompts. Specifically, the coupling function is implemented as a linear layer mapping the $d_{v}$-dimensional input to $d_{l}$, acting as a bridge between the two branches. Subsequently, these aligned visual features serve as a knowledge base for a cross-modal attention module, allowing the textual input to dynamically retrieve and integrate relevant visual semantics. Finally, the synthesized visual-guided representation is utilized as a prefix to the language branch, thereby fostering a tightly coupled cross-modal understanding. The Visual-Language prompt integration process is performed as described in \textbf{Algorithm \ref{alg1}}.

\subsection{Consistency Training}
\label{function_4}

To effectively mitigate the distribution shift between the training phase (where authentic images are available) and the inference phase (where only reconstructed images are used), we propose a joint training strategy incorporating consistency regularization. Specifically, we utilize the reconstructed image $d$, generated by the Stable Diffusion model conditioned on the input text $x$. The CLIP image encoder is employed to encode both the reconstructed image $d$ and the authentic image $a$ into their respective visual embeddings. 

For the reconstructed and authentic pathways, we independently generate visual prompts $V_{p}^{d}$ and $V_{p}^{a}$ using the VPG module. These prompts are fused with their corresponding visual embeddings via residual connections. Subsequently, a Feed-Forward Network (FFN) is applied to project the visually embedded prompts to match the dimension of the word embeddings:
\begin{equation}F^{k}=FFN(V^{k}\oplus V_{p}^{k})),\quad\text{where } k \in \{a, d\}\end{equation}
where $F^{d}\in\mathbb{R}^{1\times d}$ and $F^{a}\in\mathbb{R}^{1\times d}$ represent the projected visual representations of the reconstructed and authentic images, respectively; $d$ denotes the word embedding dimension, and $\oplus$ denotes the tensor concatenation operation. Next, $F^{d}$ and $F^{a}$ are integrated with the text prompt embedding $\hat{X}$ and fed into the Multimodal Transformer to perform the MMT task:
\begin{equation}Z^{k}=F^{k}+X_{p}^{k}\oplus X , \quad\text{where } k \in \{a, d\}\end{equation}
where $\hat{X}=X_{p}\oplus X$. Let the target sentence be denoted as $y=(y_{1},...,y_{M})$. The translation training objectives for both branches are formulated as negative log-likelihood losses:
\begin{equation}\mathcal{L}_{sdf}=-\sum_{j=1}^{M}\log p(y_{j}\mid y_{<j},x,d)\end{equation}
\begin{equation}\mathcal{L}_{aut}=-\sum_{j=1}^{M}\log p(y_{j}\mid y_{<j},x,a)\end{equation}

To enforce internal consistency between the handling of reconstructed and authentic images, and to ensure the reliability of the reconstructed path during inference, we introduce a distributional alignment loss. We employ the Kullback-Leibler (KL) divergence to minimize the discrepancy between the predicted probability distributions of the reconstructed path ($y^d$) and the authentic path ($y^a$):
\begin{equation}\mathcal{L}_{kl}=\sum_{j}^{M}f_{\theta}\big(y_{j}|y_{<j},x,d\big)\mathrm{log}\frac{f_{\theta}\big(y_{j}|y_{<j},x,d\big)}{f_{\theta}\big(y_{j}|y_{<j},x,a\big)}\end{equation}

Finally, the overall training objective is a weighted sum of the translation losses and the consistency regularization term:
\begin{equation}\mathcal{L}_{total}=\mu(\mathcal{L}_{sdf}+\mathcal{L}_{aut})+\lambda\mathcal{L}_{kl}\end{equation}
where $\mu$ and $\lambda$ are hyperparameters balancing the contributions of $\mathcal{L}_{sdf}$, $\mathcal{L}_{aut}$, and $\mathcal{L}_{kl}$. The model is optimized by minimizing $\mathcal{L}_{total}$.

\subsection{Language Translation}
\label{function_5}

\begin{figure}[htbp]
	\centering %
	\includegraphics[width=0.8\columnwidth]{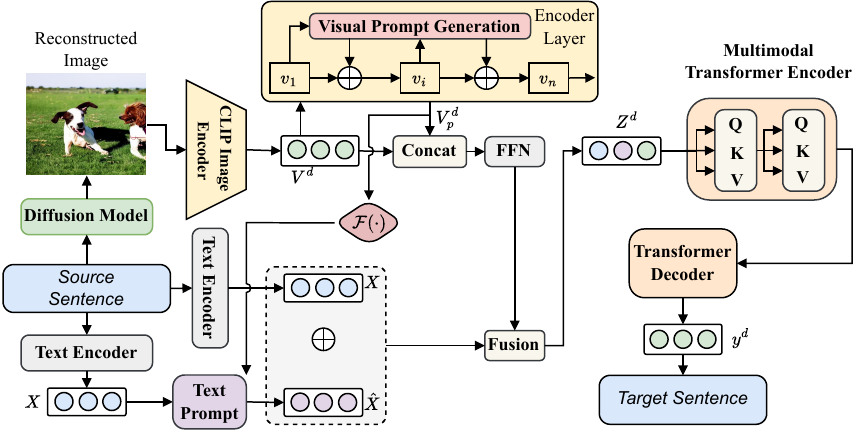}
	\caption{Our inference process uses only the source sentence and the reconstructed image as input.}
	\label{Figure 4}
\end{figure}

During the inference phase, we employ only the source sentence and its corresponding reconstructed image as inputs, thereby eliminating the dependency on authentic visual data. Specifically, we synthesize the source text embeddings, the visual representations derived from the reconstructed image, and the dual-branch prompts to construct the multimodal representation $Z^{d}$. As illustrated in Fig.~\ref{Figure 4}, $Z^{d}$ encapsulates not only the linguistic context of the source sentence but also the global and local spatial features extracted from the reconstructed visual input. 

Subsequently, $Z^{d}$ is fed into the Multimodal Transformer Encoder for hierarchical feature extraction and semantic modeling. The resulting encoded representation is then passed to the Transformer decoder, which autoregressively generates the target translation. The decoder leverages a self-attention mechanism to attend to previously generated tokens while dynamically integrating the multimodal context from the encoder to capture complex semantic dependencies. Through this iterative process, the model yields the final target translation $y^{d}$. This mechanism effectively decouples the inference process from paired authentic images, addressing a critical limitation of conventional MMT systems.

\section{Experimental Setup}

In this section, we detail the datasets, implementation configurations, and baseline models utilized in our study. We then present a comprehensive series of evaluations conducted on both the baseline and our proposed D$^{2}$P-MMT model. These experiments are designed to validate the efficacy of our approach and to rigorously assess the capability of the dual-branch prompting strategy in capturing latent semantic alignment between textual and visual modalities. All translation experiments were performed on the standard Multi30K benchmark, specifically focusing on the English $\to$ German (En-De) and English $\to$ French (En-Fr) translation tasks.

\subsection{Dataset}

We evaluate our method on the Multi30K dataset \cite{elliott2016multi30k}, the standard benchmark for Multimodal Machine Translation. Multi30K is a multilingual extension of the Flickr30K dataset \cite{young2014image}, containing images paired with English descriptions and their manual translations in German (De), French (Fr), and Czech (Cs). The dataset comprises 29,000 text-image pairs for training and 1,014 pairs for validation.

To comprehensively assess model performance, we utilize the following three test sets:
\begin{itemize}
    \item \textbf{Test2016} \cite{elliott2016multi30k}: Contains 1,000 text-image pairs from the original Multi30K distribution.
    \item \textbf{Test2017} \cite{elliott2017findings}: Comprises 1,000 text-image pairs introduced in WMT2017. This set is distinct from the training distribution and is characterized by more complex linguistic structures in the source sentences.
    \item \textbf{MSCOCO} \cite{elliott2017findings}: Includes 461 text-image pairs derived from the MSCOCO dataset. We explicitly utilize this set as an out-of-domain benchmark to evaluate model robustness. Due to significant domain shifts—including ambiguous verbs, diverse scene compositions, and non-domain-specific content—MSCOCO poses a greater challenge and rigorously tests the model's ability to generalize beyond the training distribution.
\end{itemize}

Following the preprocessing protocol~\cite{wu2021good}, we employ joint Byte Pair Encoding for subword segmentation. We perform 10,000 merge operations on the combined source and target corpora, resulting in shared vocabularies of 9,712 tokens for the English-German (En-De) task and 9,544 tokens for the English-French (En-Fr) task.

\subsection{Model Setting}
\subsubsection{Implementation Details.} 
Given the limited scale of the Multi30K dataset, prior research indicates that compact model architectures yield superior performance \cite{wu2021good}. Accordingly, we adopt the Transformer-Tiny configuration as our backbone. The translation model comprises a 4-layer encoder and a 4-layer decoder. Each layer is configured with a hidden dimension of 128, a feed-forward network intermediate dimension of 256, and 4 attention heads.
Our implementation is built upon the Fairseq framework\footnote{\url{https://github.com/facebookresearch/fairseq}}. For visual reconstruction, we utilize the \texttt{stable-diffusion-v1-4} model\footnote{\url{https://huggingface.co/CompVis/stable-diffusion-v1-4}} from the Hugging Face library. We adhere to the default generation parameters: 50 denoising steps, a guidance scale of 7.5, and a fixed random seed of 47 to ensure reproducibility. The generation batch size is set to 8. For visual feature extraction, we employ the pre-trained ViT-B/32 CLIP model\footnote{\url{https://github.com/openai/CLIP}} to process both authentic and reconstructed images.
During training, we optimize the model parameters using the Adam optimizer \cite{kinga2015method} with $\beta_{1}=0.9$ and $\beta_{2}=0.98$. The learning rate is set to $1 \times 10^{-5}$ with a warm-up period of 2000 steps. We apply a dropout rate of 0.3 and label smoothing of 0.1 for regularization. The model is trained on four NVIDIA A40 GPUs, with an effective batch size of 2048 tokens per update and a gradient accumulation frequency of 4. During inference, we select the reconstructed image most semantically aligned with the source text and generate target translations using beam search with a beam width of 5.

\subsubsection{Evaluation Metrics.} 
To ensure a fair comparison with existing baselines, we report the averaged performance of the final 10 checkpoints. We employ the 4-gram BLEU score as the primary quantitative metric to assess translation quality. BLEU measures the similarity between generated hypotheses and reference translations based on n-gram overlap, balancing adequacy and fluency. The BLEU score is calculated as:
\begin{equation}
\text{BLEU} = \text{BP} \cdot \exp \left(\sum_{n=1}^{N} w_{n} \cdot \log P_{n}\right),
\end{equation}
where $P_{n}$ denotes the n-gram precision, $N$ is the maximum n-gram order (set to 4 by default), and $w_{n} = 1/N$ represents the uniform weight for each n-gram. BP is the brevity penalty, defined as:
\begin{equation}
\text{BP} = 
\begin{cases} 
1 & \text{if } l_c > l_r \\ 
\exp\left(1 - \frac{l_r}{l_c}\right) & \text{if } l_c \leq l_r 
\end{cases},
\end{equation}
where $l_c$ is the length of the generated translation and $l_r$ is the effective reference length. All reported BLEU scores are computed on the respective test sets following standard evaluation protocols.

\subsection{Baselines}
To verify the effectiveness of the proposed model, we compared it with baseline methods from the state-of-the-art Multimodal Machine Translation tasks. The baseline methods are categorized into three types: Text-Only Transformer, Image-free systems, and Image-dependent systems. The Text-Only Transformer is implemented with the Transformer-Tiny configuration, using only text as input. The Image-free systems use only text during inference, while the Image-dependent systems use both text and images during inference.

\subsubsection{\textbf{Text-Only Transformer}}
\begin{itemize}
\item \textbf{Transformer-Tiny}~\cite{vaswani2017attention} implements a lightweight self-attention-based Transformer architecture that uses only textual input for translation..
\end{itemize}

\subsubsection{\textbf{Image-free Methods}}
\begin{itemize}
\item \textbf{UVR-NMT}~\cite{zhang2020neural} introduces a universal visual representation to replace authentic images and mitigate the scarcity of bilingual multimodal data in MMT.
\item \textbf{ImagiT}~\cite{long2021generative} proposes a generative translation method based on imagination, which constructs consistent visual representations from the source text without authentic images.
\item \textbf{IKD-MMT}~\cite{peng2022distill} adopts an inverted knowledge distillation approach to transfer multimodal knowledge from vision to the MMT model using only source text.
\item \textbf{RMMT}~\cite{wu2021good} incorporates a retrieval-based strategy that enhances translation by retrieving relevant visual information associated with the source sentence.
\end{itemize}

\subsubsection{\textbf{Image-dependent Methods}}
\begin{itemize}
\item \textbf{Doubly-ATT}~\cite{calixto2017doubly} proposes a dual-attention-based MMT framework that jointly encodes source text and visual features via two separate attention streams.
\item \textbf{DCCN}~\cite{lin2020dynamic} introduces a dynamic context-guided capsule network, which integrates visual and textual information through dynamic routing and context-aware fusion.
\item \textbf{Gumbel-Attention}~\cite{liu2021gumbel} utilizes a Gumbel-attention mechanism to generate image-aware text features, reducing interference from irrelevant visual regions.
\item \textbf{Gated Fusion}~\cite{wu2021good} designs a gated fusion module that regulates the degree of modality fusion with a gating matrix, enhancing interpretability and interaction control.
\item \textbf{Selective Attention}~\cite{li2022vision} proposes a selective attention mechanism that emphasizes stronger visual semantics to guide translation.
\item \textbf{Noise-robust}~\cite{ye2022noise} constructs a relation-aware attention module using cross-modal interaction masks to suppress noisy visual signals and enhance model robustness.
\item \textbf{VALHALLA}~\cite{li2022valhalla} employs a visual hallucination module to generate pseudo-visual features from text, enabling multimodal training without paired images.
\item \textbf{MMT-VQA}~\cite{zuo2023incorporating} incorporates Visual Question Answering (VQA) to inject question-answer supervision into the MMT process, improving visual reasoning through explicit probing.
\end{itemize}

\section{Experimental Results}

\begin{table*}[htbp]
\centering
\small
\caption{The BLEU scores for English-German and English-French translation tasks on the Multi30K test sets. Here we let $*$ represent ensembled models. ``(A)'' indicates the use of authentic images during the inference stage, while ``(R)'' represents the use of reconstructed images. Bold values indicate the highest BLEU scores, and averages are rounded to two decimal places. Some of the results are taken from the work of Li et al.\cite{li2022vision}}
\begin{tabularx}{\linewidth}{l*{7}{>{\centering\arraybackslash}X}}
    \toprule
    \multirow{2}{*}{Method} & \multicolumn{3}{c}{English-German} & \multicolumn{3}{c}{English-French} & \multirow{2}{*}{Average} \\
    \cmidrule(lr){2-4} \cmidrule(lr){5-7}
                           & Test2016 & Test2017 & MSCOCO & Test2016 & Test2017 & MSCOCO &  \\
    \midrule
    \multicolumn{8}{c}{Text-Only Transformer} \\
    \midrule
    Transformer-Tiny & 40.69  & 34.26  & 30.52  & 62.84  & 54.35  & 44.81  & 44.58  \\
    \midrule
    \multicolumn{8}{c}{Image-free Methods} \\
    \midrule
    UVR-MMT & 36.90  & 28.60  & -     & 58.30  & 48.70  & -     & - \\
    ImagiT & 38.50  & 32.10  & 28.70  & 59.70  & 52.40  & 45.30  & 42.78  \\
    IKD-MMT & 41.28  & 33.83  & 30.17  & 62.53  & 54.84  & -     & - \\
    RMMT*  & 41.40  & 32.90  & 30.00  & 62.10  & 54.40  & 44.50  & 44.22  \\
    VALHALLA* & 42.70  & 35.10  & 30.70  & 63.10  & 56.00  & \textbf{46.50}  & 45.68  \\
    \midrule
    \multicolumn{8}{c}{Image-dependent Methods} \\
    \midrule
    Doubly-ATT & 42.45  & 33.95  & 29.63  & 61.99  & 53.72  & 45.16  & 44.32  \\
    DCCN  & 39.70  & 31.00  & 26.70  & 61.20  & 54.30  & 45.40  & 43.05  \\
    Gumbel-Attention & 39.20  & 31.40  & 26.90  & -     & -     & -     & - \\
    Gated Fusion* & 41.96  & 33.59  & 29.04  & 61.69  & 54.85  & 44.86  & 44.33  \\
    Selective Attention & 41.84  & 34.32  & 30.22  & 62.24  & 54.52  & 44.82  & 44.66  \\
    Noise-robust & 42.56  & 35.09  & \textbf{31.09}  & 63.24  & 55.48 & 46.34  & 45.63  \\
    VALHALLA & 42.60  & 35.10  & 30.70  & 63.10  & 56.00  & 46.40  & 45.65  \\
    MMT-VQA & 42.55  & 34.58  & 30.96  & 62.24  & 54.89  & 45.75  & 45.16  \\
    \rowcolor{gray!20}
    D$^{2}$P-MMT (A) & 42.72  & 35.24  & 30.93  & 63.04  & 55.13  & 45.44  & 45.42  \\
    \rowcolor{gray!20}
    D$^{2}$P-MMT (R) & \textbf{43.12}  & \textbf{35.54}  & 31.01  & \textbf{63.70}  & \textbf{56.62}  & 46.23  & \textbf{46.04}  \\
    \bottomrule
\end{tabularx}
\label{table1}
\end{table*}

\subsection{Main Results}
As presented in Table~\ref{table1}, we report the quantitative results in terms of BLEU scores for all comparative models on the English-German (En-De) and English-French (En-Fr) translation tasks within the Multi30K benchmark and the out-of-domain MSCOCO test set.

Comparison with the unimodal baseline reveals that our model significantly outperforms the text-only Text-Only Transformer across all test sets, emphasizing the critical role of visual modality in enhancing NMT performance. With Transformer-Tiny as the backbone, D$^{2}$P-MMT achieves average BLEU scores of 36.56 on the English-German task and 55.52 on the English-French task. Specifically, our method yields substantial improvements of approximately \textbf{+1.40} and \textbf{+1.52} BLEU points over the text-only baseline, respectively.

In comparison with state-of-the-art MMT approaches, D$^{2}$P-MMT consistently surpasses both VALHALLA* (which generates visual features from text) and standard VALHALLA (which relies on authentic images). Notably, when compared to MMT-VQA \cite{zuo2023incorporating}, a recent method that enhances image-text modeling via visual question answering, our model achieves consistent gains of \textbf{+0.57}, \textbf{+0.96}, and \textbf{+0.05} BLEU on the En-De test sets (Test2016, Test2017, MSCOCO), and \textbf{+1.46}, \textbf{+1.73}, and \textbf{+0.48} BLEU on the corresponding En-Fr test sets. Similar performance trends are observed against other representative baselines, further validating the robustness of our framework. We attribute these improvements to the proposed dual-branch prompting strategy, which facilitates joint learning from both authentic and reconstructed images. This strategy effectively captures multi-level visual features and contextual semantic cues while mitigating adverse cross-modal interference.

Interestingly, while the translation performance using reconstructed images (D$^{2}$P-MMT(R)) is comparable to that using authentic images (D$^{2}$P-MMT(A)), the reconstructed-image-based results consistently outperform those based on authentic images. This empirical evidence suggests that the reconstructed visual representation effectively excludes extraneous visual noise found in authentic images, resulting in superior semantic alignment with the source text.

Finally, it is encouraging to note that D$^{2}$P-MMT significantly outperforms all baseline systems in both Image-Free and Image-Dependent settings, achieving state-of-the-art performance on the test sets of both translation tasks. By successfully decoupling the inference process from the reliance on authentic images, our model demonstrates a stronger capability to comprehend complex multimodal contexts, leading to more robust and accurate translation predictions.

\subsection{Model Analysis}

In this section, we provide a comprehensive analysis to further elucidate the efficacy and robustness of D$^{2}$P-MMT. Unless otherwise stated, all ablation studies and analyses are conducted on the Multi30K English-German translation task.

\begin{table}[htbp]
\centering
\caption{Number of parameters and BLEU scores of different MMT models on English-German task.}

\setlength{\arrayrulewidth}{0pt}
\begin{tabular}{@{}|lc ccc cc|@{}} 
    \toprule 
    \multirow{2}{*}{Model} & \multirow{2}{*}{\#Params} & \multicolumn{3}{c}{English-German} & \multirow{2}{*}{Average} & \multirow{2}{*}{\#Speed(Tokens/s)}  \\
    \cmidrule(lr){3-5} 
                       &                           & Test2016 & Test2017 & MSCOCO &  \\
    \midrule
    \multicolumn{7}{c}{Text-Only Transformer} \\
    \midrule
    Transformer-Tiny      & 2.60M  & 40.69 & 34.26 & 30.52  & 35.16  & 1798.51 \\
    Transformer-Small     & 36.5M  & 39.68 & 32.99 & 28.50  & 33.72  & 1662.10 \\
    Transformer-Base      & 49.1M  & 38.33 & 31.36 & 27.54  & 32.41  & 996.30 \\
    \midrule
    \multicolumn{6}{c}{Our proposed method} \\
    \midrule
    D$^{2}$P-MMT(A)-Tiny  & 4.35M  & 42.72 & 35.24 & 30.93 & 36.30 & 1971.23 \\
    D$^{2}$P-MMT(A)-Small & 36.85M & 42.18 & 34.13 & 31.49 & 35.93 & 1795.95 \\
    D$^{2}$P-MMT(A)-Base  & 49.25M & 41.78 & 34.49 & 30.37 & 35.55 & 1379.56 \\
    \rowcolor{gray!20}
    D$^{2}$P-MMT(R)-Tiny  & 4.35M  & \textbf{43.12} & \textbf{35.54} & 31.01 & \textbf{36.56}  & \textbf{1971.23} \\
    D$^{2}$P-MMT(R)-Small & 36.85M & 42.06 & 34.38 & \textbf{31.78}& 36.07 & 1795.95 \\
    D$^{2}$P-MMT(R)-Base  & 49.25M & 41.82 & 34.45 & 30.51 & 35.59 & 1379.56 \\
    \bottomrule 
    \end{tabular}
\label{table2}
\end{table}

\subsubsection{Generalization to Different Model Architectures}

To validate the generalization capability of our proposed framework, we extend our experiments to various Transformer configurations, maintaining consistent hyperparameters with the primary Multimodal Transformer setup. Furthermore, we address the often-overlooked aspect of model complexity, which is particularly critical given the limited scale of standard MMT benchmarks. As highlighted by Wu et al. \cite{wu2021good}, larger architectures such as Transformer-Base and Transformer-Small are prone to overfitting on small-scale datasets, whereas compact models like Transformer-Tiny often exhibit superior robustness and efficiency.

The experimental results, presented in Table~\ref{table2}, corroborate these observations. D$^{2}$P-MMT maintains an optimal balance in parameter efficiency, effectively mitigating the risk of overfitting. Although our model introduces a marginal increase in parameters compared to the unimodal Transformer-Tiny baseline, it yields exceptional performance gains. This underscores the critical trade-off between parameter count and model capacity in Transformer-based systems, especially within the data-constrained domain of Multimodal Machine Translation. In addition to parameter efficiency, Table~\ref{table2} also lists the processing times for various methods. Although our model is slower than the plain text Transformer-Tiny model due to additional visual processing, the speed disadvantage is not significant. Meanwhile, our model achieves a better balance between processing speed and translation quality, with the improvement in translation quality more than offsetting the additional computational overhead.

\begin{table}[htbp]
\centering
\caption{BLEU scores of our method and other regularization methods.}
    \setlength{\arrayrulewidth}{0pt}
    \begin{tabular}{@{}|l ccc c|@{}}
    \toprule
    \multirow{2}{*}{Model} & \multicolumn{3}{c}{English-German} & \multirow{2}{*}{Average} \\
    \cmidrule(lr){2-4} 
                       & Test2016 & Test2017 & MSCOCO &  \\
    \midrule
    Noise                  & 41.86          & \textbf{35.72} & 30.77          & 36.12          \\
    \rowcolor{gray!20}
    D$^{2}$P-MMT(R)        & \textbf{43.12} & 35.54          & \textbf{31.01} & \textbf{36.56} \\
    \bottomrule
    \end{tabular}
\label{table3}
\end{table}

\begin{figure}[t]
  \centering
  \includegraphics[width=0.8\textwidth]{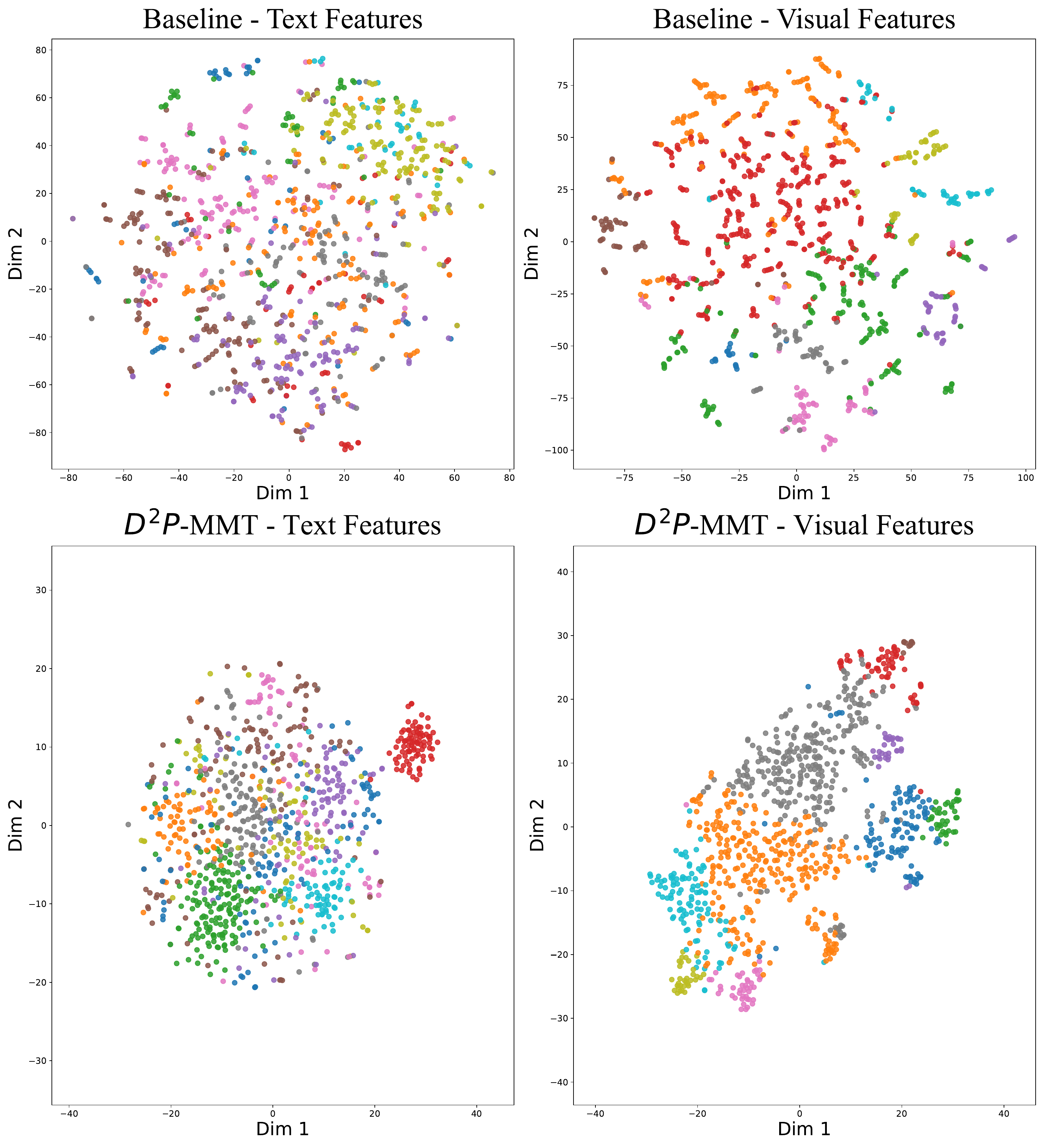}
  \caption{t-SNE visualization of unimodal feature representations for 1,000 samples from the Multi30K dataset. The top row shows features from a baseline configuration, while the bottom row displays the representations enhanced by our method.}
  \label{Figure 5} 
\end{figure}

\subsubsection{Impact of Visual Semantic Relevance}

Following established protocols for probing visual dependency in MMT \cite{caglayan2019probing}, we employ adversarial settings to evaluate the sensitivity of our model to non-corresponding visual contexts. Specifically, we substitute aligned visual features with random noise vectors \cite{zhang2025reliable} to simulate a complete loss of visual semantic information, a setting we denote as \textbf{Noise}. We then compare the performance degradation relative to our standard D$^{2}$P-MMT.

As presented in Table~\ref{table3}, D$^{2}$P-MMT significantly outperforms the Noise baseline, achieving an average BLEU improvement of \textbf{+0.44} across the three English-German test sets. This performance gap underscores two critical findings: first, the reconstructed images serve as highly effective visual contexts by eliminating redundancy found in authentic images; and second, our dual-branch framework successfully leverages this refined visual information to enhance translation accuracy, confirming the model's genuine reliance on relevant visual semantics rather than mere statistical artifacts.

\subsubsection{Impact of Dual-Branch Prompting}

To empirically assess the impact of our dual-branch prompting strategy on feature representation learning, we visualize the latent spaces of both textual and visual modalities. We employ t-SNE \cite{maaten2008visualizing} to project the high-dimensional features from a standard baseline configuration and our proposed D$^{2}$P-MMT into a 2D space, as illustrated in Fig.~\ref{Figure 5}.

A comparative analysis reveals distinct differences in feature distribution. In the baseline configuration, both textual and visual features exhibit a scattered distribution with limited separability, indicating weak semantic clustering. In sharp contrast, the representations refined by our dual-branch prompting mechanism demonstrate significantly enhanced structure. This improvement is particularly pronounced in the visual modality, where we observe sharper cluster boundaries and substantially improved intra-class compactness. These visualizations provide compelling evidence that our strategy effectively regularizes the multimodal feature space, enabling the model to not only grasp global semantic contexts but also capture fine-grained local details with greater precision.

\subsection{Ablation Study}

To provide a comprehensive evaluation of the D$^2$P-MMT framework and dissect the contribution of each individual component, we conduct a series of ablation studies. Specifically, we investigate the efficacy of leveraging reconstructed visual information and the impact of our cross-modal interaction mechanisms. Our analysis focuses on four critical aspects: (1) the contribution of the distributional alignment loss ($\mathcal{L}_{kl}$); (2) the design of the visual prompt generation strategy; (3) the role of the cross-branch coupling function; and (4) the overall effectiveness of the dual-branch prompting architecture. The quantitative results corresponding to these configurations are summarized in Table~\ref{table4}.

\begin{table}[htbp]
\centering
\caption{Ablation study over different components of our model on the English-German translation task. All results use BLEU scores.}
\setlength{\arrayrulewidth}{0pt} 
\begin{tabular}{@{}|l ccc c|@{}} 
    \toprule
    \multirow{2}{*}{Model} & \multicolumn{3}{c}{English-German} & \multirow{2}{*}{Average} \\
    \cmidrule(lr){2-4} 
                       & Test2016 & Test2017 & MSCOCO &  \\
    \midrule
    \rowcolor{gray!20}
    D$^{2}$P-MMT(R)             & \textbf{43.12} & 35.54          & 31.01          & \textbf{36.56} \\
    D$^{2}$P-MMT(A)             & 42.72          & 35.24          & 30.93          & 36.30          \\
    w/o $\mathcal{L}_{kl}$ & 41.83          & 33.94          & 29.59          & 35.12          \\
    w/o $VPG_{\text{global}}$   & 42.80          & \textbf{36.14} & 30.30          & 36.41          \\
    w/o $VPG_{\text{local}}$    & 42.29          & 34.76          & 30.63          & 35.89          \\
    w/o $\mathcal{F}(\cdot)$    & 42.63          & 35.93          & 30.12          & 36.23          \\
    w/o VPG                     & 42.57          & 35.08          & 30.88          & 36.18          \\
    w/o prompt                  & 42.36          & 35.20          & \textbf{31.60} & 36.39          \\
    \bottomrule
\end{tabular}
\label{table4}
\end{table}
\subsubsection{Contribution of the Distributional Alignment Loss ($\mathcal{L}_{kl}$)}

In this variant, we omitted the distributional alignment loss ($\mathcal{L}_{kl}$) during training to assess its specific contribution to enforcing consistency between the reconstructed and authentic inference paths. As reported in Table~\ref{table4}, the removal of $\mathcal{L}_{kl}$ leads to a significant degradation in performance. This finding underscores the critical role of this loss in bridging the distributional gap between the two visual pathways. Specifically, $\mathcal{L}_{kl}$ ensures that the target translations generated from reconstructed images remain semantically consistent with those derived from authentic images, thereby stabilizing the inference process.

Furthermore, we investigated the efficacy of alternative divergence metrics by replacing the KL divergence with Jensen-Shannon (JS) divergence, Earth Mover's Distance (EMD) \cite{rubner2000earth}, and Cosine Embedding loss, while maintaining identical hyperparameters. As shown in Table~\ref{table5}, these alternatives yielded inferior BLEU scores compared to the KL divergence. This comparative analysis suggests that KL divergence is particularly effective in mitigating the specific discrepancies between the visual prompt distributions of reconstructed and authentic images, thereby reducing the semantic representational gap between the modalities.

\begin{table}[htbp]
\centering
\caption{BLEU scores of our method and other loss strategies.}
\setlength{\arrayrulewidth}{0pt} 
\begin{tabular}{@{}|l ccc c|@{}} 
    \toprule
    \multirow{2}{*}{Model} & \multicolumn{3}{c}{English-German} & \multirow{2}{*}{Average} \\
    \cmidrule(lr){2-4} 
                       & Test2016 & Test2017 & MSCOCO &  \\
    \midrule
    JS                     & 41.67          & 34.84          & 30.74          & 35.75          \\
    EMD                    & 40.99          & 33.94          & \textbf{31.61} & 35.51          \\
    Cosine                 & 40.98          & 34.37          & 29.45          & 34.93          \\
    \rowcolor{gray!20}
    Ours(R)                & \textbf{43.12} & \textbf{35.54} & 31.01          & \textbf{36.56} \\
    \bottomrule
    \end{tabular}
\label{table5}
\end{table}

\subsubsection{Design of the Visual Prompt Generation Strategy}

The Visual Prompt Generation (VPG) module is designed with a dual-branch architecture to capture both global context and local details. To validate this design, we conducted ablation studies by independently removing the global processing branch (w/o $VPG_{\text{global}}$) and the local processing branch (w/o $VPG_{\text{local}}$). 

As shown in Table~\ref{table4}, relying on a single branch results in a noticeable decline in translation performance. The global branch provides a holistic scene understanding, while the local branch captures fine-grained details; their synergistic combination allows the model to form a comprehensive visual representation that is robust to variations in the input image.

To further validate the contributions of the $VPG$ module, we conducted a detailed ablation study of its internal architecture, as shown in Table~\ref{table6}. These experiments demonstrate that relying solely on uniform standard ${3 \times 3}$ convolutions is insufficient to simultaneously capture both macroscopic scene context and microscopic object details. Moreover, simply increasing the receptive field in the global branch or using uniform ${3 \times 3}$ convolutions in the local branch limits the model's ability to extract rich semantics from images. In contrast, our proposed hybrid architecture—combining a global bottleneck branch with a mixed-kernel local branch—achieves the best performance. 

\begin{table}[htbp]
\centering
\caption{Impact of different receptive field strategies for visual prompt modeling on translation performance. Notation format: ``Global Branch Config + Local Branch Config''. $M1$ and $M2$ denote replacing all kernels in the corresponding branch with uniform $\text{Conv2d}_{3 \times 3}$ and $\text{Conv2d}_{5 \times 5}$, respectively.}
\setlength{\arrayrulewidth}{0pt} 
\begin{tabular}{@{}|l ccc c|@{}} 
    \toprule
    \multirow{2}{*}{Modules} & \multicolumn{3}{c}{English-German} & \multirow{2}{*}{Average} \\
    \cmidrule(lr){2-4} 
                       & Test2016 & Test2017 & MSCOCO &  \\
    \midrule
    $\text{M1+M1}$ & 42.04  & 35.28  & 29.51  & 35.61  \\
    $VPG_{\text{global}}+ M1$ & 42.38  & \textbf{35.76}  & 29.84  & 35.99  \\
    $VPG_{\text{local}}+ M2$ & 42.12  & 34.34  & 30.50  & 35.65  \\
    \rowcolor{gray!20}
    Ours(R)                & \textbf{43.12} & 35.54 & \textbf{31.01} & \textbf{36.56} \\
    \bottomrule
    \end{tabular}
\label{table6}
\end{table}

\subsubsection{Role of the Cross-Branch Coupling Function}

As detailed in Section~\ref{function_vl}, D$^2$P-MMT employs a linear coupling function $\mathcal{F}(\cdot)$ to explicitly condition textual prompts on visual prompts ($V_{p} \to X_{p}$). To evaluate the specific role of this component, we compared it against an alternative implementation using a one-dimensional (1D) convolution. The results in Table~\ref{table7} demonstrate the superiority of our linear projection approach. While 1D convolutions effectively extract local sequential features, our linear design better preserves global feature relationships in high-dimensional spaces with minimal information loss. This suggests that the coupling function plays a pivotal role in maintaining model expressiveness and facilitating direct cross-modal alignment.

\begin{table}[htbp]
\centering
\caption{Results of different coupling strategies on the Multi30K English-German test set.}
\setlength{\arrayrulewidth}{0pt} 
\begin{tabular}{@{}|l ccc c|@{}} 
    \toprule
    \multirow{2}{*}{Model} & \multicolumn{3}{c}{English-German} & \multirow{2}{*}{Average} \\
    \cmidrule(lr){2-4} 
                       & Test2016 & Test2017 & MSCOCO &  \\
    \midrule
    Conv1d                 & 42.12          & 34.74          & 30.33          & 35.73          \\
    \rowcolor{gray!20}
    Ours(R)                & \textbf{43.12} & \textbf{35.54} & \textbf{31.01} & \textbf{36.56} \\
    \bottomrule
    \end{tabular}
\label{table7}
\end{table}

\subsubsection{Overall Effectiveness of the Dual-Branch Prompting Architecture}

Finally, we evaluate the overall effectiveness of the dual-branch prompting framework. First, we analyze the necessity of the projection mechanism by removing the coupling function and directly embedding visual prompts into the language branch (denoted as ``w/o $\mathcal{F}(\cdot)$''). The significant performance drop observed in Table~\ref{table4} highlights the crucial role of the coupling function in harmonizing the feature spaces of the two modalities.

Second, we compare our approach with an independent prompting strategy \cite{zhang2024dept}. When visual prompts are entirely removed (w/o $VPG$), D$^2$P-MMT suffers performance decreases of 0.55, 0.46, and 0.13 BLEU on the three Multi30K test sets, respectively. Furthermore, completely removing the dual-branch prompting mechanism (w/o $prompt$) results in an average drop of 0.17 BLEU on the English-German task. These results collectively demonstrate that dual-branch prompting facilitates deep, structured interaction between modalities. By effectively integrating visual and textual features, the framework significantly enriches contextual awareness and enhances overall translation quality.

\subsection{Case Study}

To provide qualitative insights into the efficacy of our approach, we present a comparative analysis of translation examples generated by different systems, as illustrated in Table~\ref{table8}.

\begin{table}[htbp]
\centering
\caption{Two translation cases of three systems on the English-German task. The red and blue highlight error and correct translations respectively.}
\resizebox{1\linewidth}{!}{
\begin{tabular}{cll}
    \toprule
    \multirow{5}{*}{\resizebox{0.2\linewidth}{!}{\includegraphics{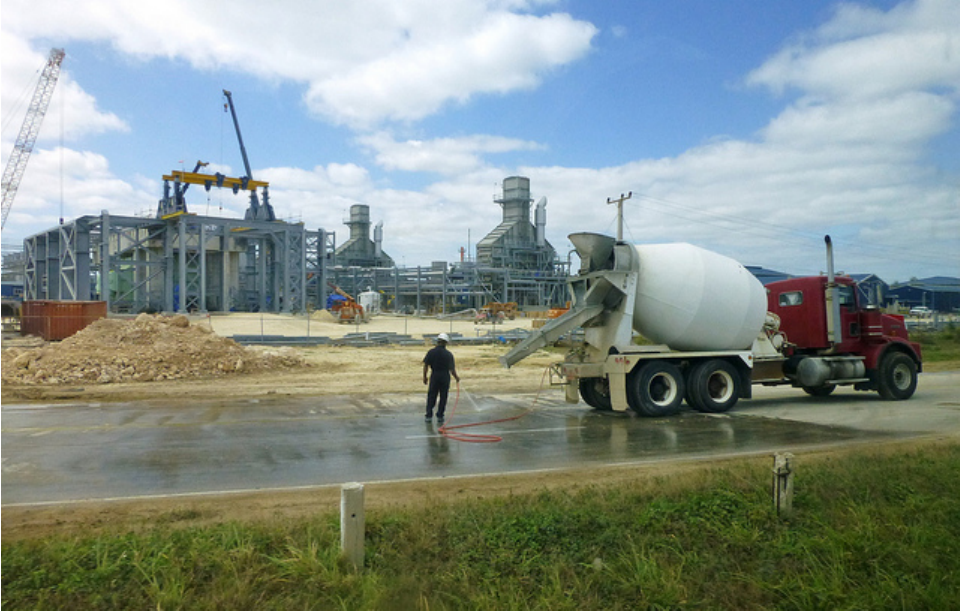}}} &
    SRC:  & a cement truck pours fresh cement on the road. \\
          & REF:  & ein zementlaster gießt \textcolor[HTML]{21AEE7}{frischen} zement \textcolor[HTML]{21AEE7}{auf die straße} . \\
          & Text-Only:  & \textcolor{red}{arbeiter arbeiten an einer straße}. \\
          & MMT-VQA:  & ein zementlaster gießt \textcolor{red}{frische} zement \textcolor{red}{auf den boden} . \\
          & D$^{2}$P-MMT:  & ein zementlaster gießt \textcolor[HTML]{21AEE7}{frischen }zement \textcolor[HTML]{21AEE7}{auf die straße}. \\
    \midrule
    \multirow{5}[2]{*}{\resizebox{0.2\linewidth}{!}{\includegraphics{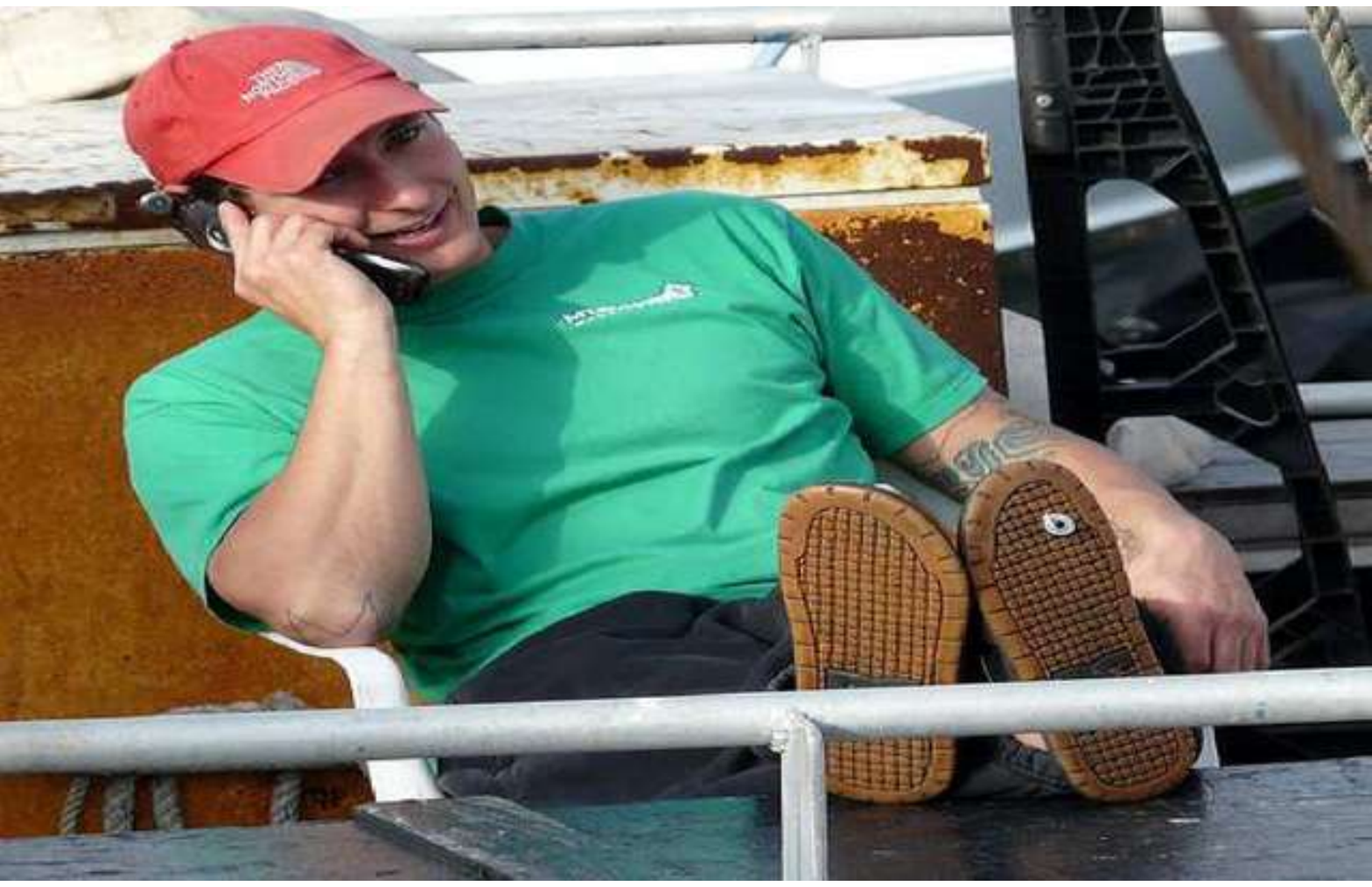}}} & 
    SRC:  & a man talks on the phone with his feet up . \\
          & REF:  & ein mann telefoniert mit \textcolor[HTML]{21AEE7}{hochgelegten} füßen . \\
          & Text-Only:  & ein mann \textcolor{red}{sitzt auf einer bank} und telefoniert . \\
          & MMT-VQA:  & ein mann telefoniert mit \textcolor{red}{den} füßen . \\
          & D$^{2}$P-MMT:  & ein mann telefoniert mit \textcolor[HTML]{21AEE7}{hochgelegten} füßen . \\
    \bottomrule
    \end{tabular}
}
\label{table8}
\end{table}

In the first example, the baseline system fails to grasp the visual essence of the scene, erroneously translating the description as ``arbeiter arbeiten an einer Straße'' (\textit{Workers are working on a road}). The second comparative system misinterprets the spatial context, rendering ``on the street'' as ``auf den Boden'' (\textit{on the ground}), resulting in semantic inaccuracy. In sharp contrast, our D$^2$P-MMT accurately captures fine-grained details, correctly translating the key phrases ``frischen'' (fresh) and ``auf die Straße'' (\textit{on the street}), thereby preserving the original semantic nuance.

In the second example, the competing MMT system overlooks the directional preposition ``up'' within the German context, failing to accurately convey the state of the foot. Conversely, our system correctly generates the German term ``hochgelegten'', effectively encapsulating the concept of a ``raised foot''. These qualitative results demonstrate that by incorporating reconstructed visual information, our model significantly enhances the capability to resolve lexical and visual ambiguities. This further corroborates that the dual-branch prompt enhancement strategy effectively grounds translations in local visual details, leading to more precise and contextually aware outputs.

\begin{figure}[htbp]
  \centering
  \includegraphics[width=0.7\textwidth]{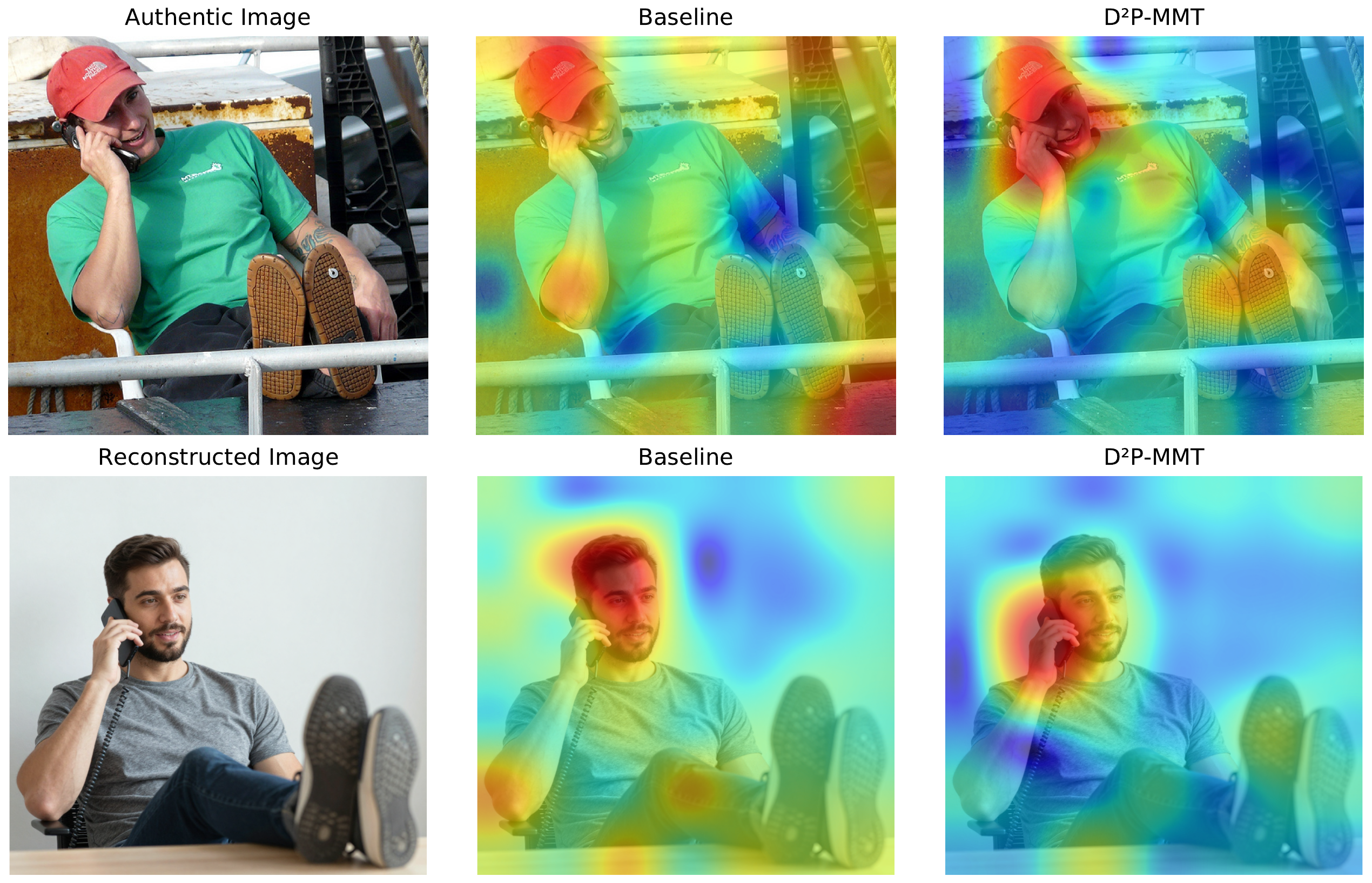}
  \caption{Noise-Robust Attention Heatmaps.}
  \label{Figure 6} 
\end{figure}

To further investigate the model’s behavior in semantically ambiguous contexts, we visualized the attention heatmap distribution of the second example across different models. As illustrated in Fig.~\ref{Figure 6}, the reconstructed images provide cleaner visual cues. While the baseline model tends to fixate on visually salient but text-irrelevant background regions, D$^{2}$P-MMT consistently directs its attention to text-aligned visual features, such as the raised feet and the phone. This focused attention pattern demonstrates that our diffusion-based reconstruction effectively filters out extraneous visual noise, enabling the model to ground translations in semantically meaningful regions.

\section{Conclusion}

In this work, we propose D$^{2}$P-MMT, a robust multimodal machine translation framework that addresses visual redundancy and weak cross-modal alignment by leveraging diffusion-based image reconstruction and a dual-branch prompting strategy. The reconstruction mechanism effectively filters irrelevant visual noise, while the prompting design enables deeper and more precise interactions between textual and visual representations, decoupling inference from strict reliance on paired authentic images. Experimental results on the Multi30K benchmark demonstrate state-of-the-art performance in both Image-Free and Image-Dependent settings, validating the effectiveness of the proposed approach. Future work will aim to generalize this framework to datasets with diverse vocabularies and to further optimize the efficiency and adaptability of the model, paving the way for more robust and scalable unsupervised multimodal machine translation.

\begin{acks}
This work was supported by the National Natural Science Foundation of China (Grant No. 62506311 and 62506310), the Sichuan Science and Technology Program (No. 2026NSFSC1481 and 2026NSFSC1479), the China Postdoctoral Science Foundation under Grant Number 2025M771637 and 2025M781517, the Postdoctoral Fellowship Program of CPSF under Grant Number GZB20250417, the Fundamental and Interdisciplinary Disciplines Breakthrough Plan of the Ministry of Education of China (Grant No. JYB2025XDXM211) and the Fundamental Research Funds for the Central Universities (Grant No. 2682025CX105).
\end{acks}

\bibliographystyle{ACM-Reference-Format}
\bibliography{main}

\end{document}